# Estimating Parameters of the Tree Root in Heterogeneous Soil Environments via Mask-Guided Multi-Polarimetric Integration Neural Network

Hai-Han Sun, Yee Hui Lee, Qiqi Dai, Chongyi Li, Genevieve Ow,
Mohamed Lokman Mohd Yusof, and Abdulkadir C. Yucel

*Abstract*— Ground-penetrating radar (GPR) has been used as a non-destructive tool for tree root inspection. Estimating root-related parameters from GPR radargrams greatly facilitates root health monitoring and imaging. However, the task of estimating root-related parameters is challenging as the root reflection is a complex function of multiple root parameters and root orientations. Existing methods can only estimate a single root parameter at a time without considering the influence of other parameters and root orientations, resulting in limited estimation accuracy under different root conditions. In addition, soil heterogeneity introduces clutter in GPR radargrams, making the data processing and interpretation even harder. To address these issues, a novel neural network architecture, called mask-guided multi-polarimetric integration neural network (MMI-Net), is proposed to automatically and simultaneously estimate multiple root-related parameters in heterogeneous soil environments. The MMI-Net includes two sub-networks: a MaskNet that predicts a mask to highlight the root reflection area to eliminate interfering environmental clutter, and a ParaNet that uses the predicted mask as guidance to integrate, extract, and emphasize informative features in multi-polarimetric radargrams for accurate estimation of five key root-related parameters. The parameters include the root depth, diameter, relative permittivity, horizontal and vertical orientation angles. Experimental results demonstrate that the proposed MMI-Net achieves high estimation accuracy in these root-related parameters. This is the first work that takes the combined contributions of root parameters and spatial orientations into account and simultaneously estimates multiple root-related parameters. The data and code implemented in the paper can be found at https://haihan-sun.github.io/GPR.html.

## I. Introduction

TREE root system is a key component of trees as it anchors the tree, provides support, and transports the water and nutrients necessary for trees' growth [1]. Ground-penetrating radar (GPR) as a non-destructive tool has been used for the inspection of root systems [2], [3]. Extraction of characteristic parameters of roots from GPR radargrams is an active research area as it facilitates the investigation of roots' ecological function, monitoring of root health, and reconstruction of root system architectures [3], [4]. The task, however, is challenging since the signature of reflection from a root in GPR radargrams is a complicated function of multiple root-related parameters such as the diameter, permittivity, depth, and orientations. Moreover, environmental clutter introduced by soil heterogeneity makes the processing and interpretation of GPR radargrams even more challenging [5].

Over the years, many attempts have been made to predict root-related parameters from GPR radargrams, including root diameter [6]-[12], water content [13], and biomass [14]-[16]. Traditional methods use regression models to build statistical relationships between GPR waveform parameters such as the maximum magnitude and time span with a specific root-related parameter [6]-[16]. However, several limitations exist in the traditional methods.

1) Most models are established based on the assumption that the waveform parameters are governed by a single root-related parameter, whereas the reflection from a root is affected by the joint contributions of multiple parameters in reality. The inadequate consideration of the effects from other root parameters in the models can result in high uncertainty of estimation accuracy across different survey sites. To mitigate this issue, a non-linear regression model which takes the joint effects of two root parameters on GPR signal strength into account was built in [12]. As a result, the model's estimation accuracy for root water content was significantly improved. It is easy to foresee that considering more root parameters in the estimation process could further improve the estimation accuracy.

2) Most of the parameter estimation models only consider the situation where the root's horizontal orientation is perpendicular to the scan trace and the root has no inclination angle. Such an assumption is invalid in the real situation where roots grow along different spatial directions [2], [17]. It is hard to identify roots that have complex angles with respect to the direction of survey lines as the characteristic hyperbolic reflection shape is lost [17], which adds significant uncertainty to the GPR data interpretation and parameter estimation process [5]-[7]. Specifically, as stated in [7], the existing models cannot hold their validity with unknown orientation. Therefore, further investigation should be carried out to link the characteristics of radargrams with root parameters while taking root orientations into account. Besides, knowing the root orientation also facilitates the mapping of root system architecture [4].

Another challenge for the root parameter estimation is the interference of environmental noises and clutters from the intrinsic heterogeneity of soil. This introduces difficulties in interpreting radargrams and jeopardizes the estimation accuracy [5], [6]. To eliminate the detrimental effect of the environmental clutters for analyzing GPR radargrams, it is feasible to first identify or extract the reflection signatures from GPR radargrams [18], [19]. Strategies have been proposed for automatic segmentation of reflection signatures to narrow down the region of interest (RoI) for subsequent processing in a complex GPR image. The clustering algorithms [20]-[22] were proposed to separate the hyperbolic regions in a complex GPR image by connecting pixels in adjacent columns with the same row number. Deep learning algorithms such as Viola-Jones algorithm [23], Faster Region-based Convolutional Neural Network (Faster R-CNN) [24], and Mask Region-based Convolutional Neural Network (Mask R-CNN) [25] have been used in [19], [22], [26] to detect and segment object signatures from B-scans.

Inspired by the powerful feature extraction and non-linear modeling capabilities of deep neural networks [27]-[30], a specialized neural network, called Mask-Guided Multi-Polarimetric Integration Neural Network (MMI-Net), is proposed to estimate the root-related parameters in heterogeneous soil environment using multi-polarimetric radargrams. The root-related parameters studied in this work are the depth, diameter, relative permittivity, and horizontal and vertical orientation angles. The proposed network consists of two sub-networks. One is the MaskNet that takes multi-polarimetric radargrams as inputs and automatically predicts a mask on the root reflection signature. The predicted mask highlights the root reflection region, which helps to eliminate most of the background noises. Another sub-network is the ParaNet, which takes both the masked and original multi-polarimetric radargrams as inputs and extracts the features of the root reflection pattern to perform parameter estimation. The masked images guide the ParaNet to pay more attention to the RoI, while the original images allow the network to take useful information in the background environment into account for accurate parameter estimation. A channel-attention (CA) module is introduced in the ParaNet to model interdependencies between multi-polarimetric information and emphasize informative features related to root parameters. Two sub-networks are coupled for accurate parameter estimation. The effectiveness and the advantage of the proposed network are demonstrated by experiments using both the simulated and measured GPR data and a series of comparative studies with existing work. The contributions of this work are three-fold:

1) To the best of our knowledge, the proposed method is the first that takes both the root horizontal and vertical orientations into account while estimating other root parameters. This addresses the uncertainty of root identification and parameter estimation when the root direction is unknown in the existing methods.

2) The proposed MMI-Net has a novel neural network architecture that consists of a RoI identification sub-network (MaskNet) and a parameter estimation sub-network (ParaNet). It is specially designed to alleviate the adverse effects of environmental clutter on estimating root-related parameters under heterogeneous soil conditions.

3) Compared with the existing works that only estimate a single root parameter at a time, the proposed MMI-Net achieves simultaneous estimation of multiple root-related parameters by effectively building the non-linear relationship between multi-polarimetric radargrams and root-related parameters. The success of simultaneously estimating multiple root-related parameters will facilitate further root investigations, such as the health examination and 3D reconstruction.

The rest of the paper is organized as follows. The influences of different root parameters on the multi-polarimetric radargrams are analyzed in Section II. The structure of the MMI-Net for the estimation of root parameters using multi-polarimetric radargrams is described in Section III. The implementation details of the MMI-Net and the experimental results using the simulated GPR dataset are presented in Section IV. The performance of the MMI-Net on the measured GPR dataset is examined in Section V. Comparison studies and suggestions for future improvements are discussed in Section VI. Finally, the conclusions are drawn in Section VII.

## II. CHARACTERISTICS OF MULTI-POLARIMETRIC RADARGRAMS WITH DIFFERENT ROOT PARAMETERS

As the root is an elongated object which depolarizes electromagnetic (EM) waves based on its orientation [31], in this work, a multi-polarimetric GPR is implemented to improve the detectability of tree roots [32]-[35]. A root with different parameters gives different reflection responses to multi-polarimetric components. The effects of root parameters and soil heterogeneity on the multi-polarimetric radargrams are examined in this section, which inspires the design of the proposed MMI-Net as presented in Section III.

*A. Simulation Scenario*

In the numerical study, gprMax [36] is used to obtain the responses of different polarimetric components when detecting a root with different parameters. The simulation scenario is shown in Fig. 1. A tree root with a diameter $dm$ and a relative permittivity $\varepsilon_r$ is buried in the soil. The distance from the root center to the soil surface is $d$. The root is oriented at a horizontal angle $\varphi$ and a vertical angle $\theta$, where $\varphi$ is the angle from the $x$-axis to the $y$-axis in the clockwise direction on the $xy$-plane and $\theta$ is the angle from $x$-axis to the $-z$-axis in the anti-clockwise direction. Since a root often extends to a long distance, the simulated root has its two ends extended beyond the boundary of the simulation domain. Orthogonally polarized ($x$- and $y$-polarized) sources and probes are used as transmitters and receivers to detect the object. They are separated by 10 cm and placed above the soil surface at the height of 2 cm. The excitation waveform is a Ricker pulse with a center frequency of 1 GHz and an amplitude of 1 A. The scanning is carried out by moving the source and probe along $x$-direction with a step size of 2.5 cm. At each scanning position, a scattering matrix with four polarimetric components is obtained as

$$S(t) = \begin{bmatrix} S_{xx}(t) & S_{xy}(t) \\ S_{yx}(t) & S_{yy}(t) \end{bmatrix}, \quad (1)$$

where the first and second subscripts in the matrix components represent the polarization of probe and source, respectively. 33 A-scans along the trace are collected and then combined into a B-scan radargram for each polarimetric component.

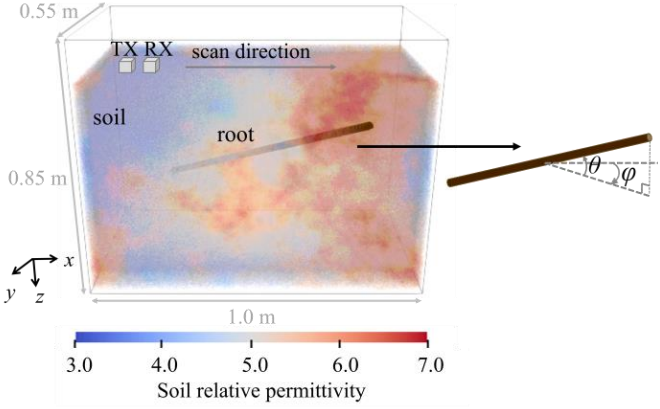

Fig. 1. Illustration of the simulation scenario. Different colors in the soil region represent different soil permittivity.

*B. Influence of Root Parameters and Soil Heterogeneity on the Radargrams of Different Polarization Components*

In this subsection, the influence of root parameters on the multi-polarimetric radargrams in an ideal homogeneous soil environment is firstly analyzed, followed by the examination of the influence of soil heterogeneity on the radargrams. The homogeneous soil has a relative permittivity of 4.9, a conductivity of 0.002 S/m, and a relative permeability of 1. These values are the average values of the heterogeneous soil in this study. When analyzing the influence of each parameter, other parameters and soil properties are fixed to the default values. The default values of the parameters are $d = 0.3\ m$, $dm = 0.04\ m$, $\varepsilon_r = 22$, $\varphi = 90°$, and $\theta = 0°$, unless otherwise specified.

**Depth $d$.** Three cases with root depth of 0.2 m, 0.3 m, and 0.4 m are simulated, and the resulting multi-polarimetric radargrams are shown in Fig. 2(a). As the root is oriented parallel to the y-axis, the $S_{yy}$ component collects the reflected signal with the largest magnitude. The $S_{xx}$ component collects the object reflection with a smaller magnitude as the root has a diameter of 4 cm, comparable to the wavelength at 1 GHz in soil. The $S_{xy}$ and $S_{yx}$ components cannot receive the object reflection as cross-polarization cannot detect elongated objects oriented parallel to one of the polarizations [37]. This also applies for Figs. 2(b) and 2(c) in situations with different diameters and relative permittivities.

As shown in Fig. 2(a), the depth mainly affects the position of the reflection signature in the time direction, which is the vertical direction in the radargrams. In addition, the depth also affects the shape of the hyperbolic signature and the intensity of the reflected signal. This is due to the spatial relationship between the data collection point and the root position, as well as the signal attenuation effect during the transmission of EM waves in the soil. A deeper position leads to a broader hyperbolic signature and a weaker signal strength. Therefore, with the prior soil properties, root depth information is carried in the spatial location, shape, and pixel values of the root reflection signature in the multi-polarimetric B-scans.

**Diameter $dm$.** Multi-polarimetric radargrams with root diameter of 2 cm, 4 cm, and 6 cm are shown in Fig. 2(b). A root with a larger diameter leads to a broader hyperbola shape with a larger signal strength, which is evident in both $S_{xx}$ and $S_{yy}$ components. Here it is noted that when the root is very thin (i.e., its diameter is not comparable to wavelength), $S_{xx}$ may not be able to collect reflected signals as the x-polarization is orthogonal to the root orientation, but $S_{yy}$ can still collect signals. Also, a root with a larger diameter has distinguishable double reflections from its top and bottom surfaces, and the time delay between the two reflections is correlated to the diameter. Therefore, with the prior soil properties, the information of a root diameter is carried in shape, pixel values, and occupied spatial area of the root reflection signature in the radargrams.

**Relative permittivity $\varepsilon_r$.** Fig. 2(c) shows the multi-polarimetric radargrams with root relative permittivity of 15, 22, and 30. The root permittivity is related to the root water content, which is a valuable parameter to examine the root health condition [38]. As shown in Fig. 2(c), higher root relative permittivity leads to a larger permittivity contrast between the root and soil, producing a larger reflected signal strength of the root. In addition, higher root relative permittivity reduces the travel speed of EM waves in the root, resulting in a larger time delay between top and bottom reflections.

**Horizontal orientation angle $\varphi$.** Fig. 2(d) shows the multi-polarimetric radargrams of a root with different horizontal orientation angles $\varphi$. The influence of $\varphi$ can be summarized in two aspects.

1) Reflected signal strength. When $\varphi = 0°$, $S_{xx}$ collects the the reflection with the maximum strength as the root direction is parallel to the x-direction. Root reflection collected by $S_{yy}$ is also distinguishable as the root diameter is large to be detected by the polarization that is orthogonal to the root orientation. The cross-polarized components, $S_{xy}$ and $S_{yx}$, cannot receive root reflection. When $\varphi$ increases from 0° to 45°, the strength of the signal collected by $S_{xx}$ decreases, while the strengths of the signals collected by $S_{xy}$, $S_{yx}$, and $S_{yy}$ increase. When $\varphi$ further increases from 45° to 90°, $S_{yy}$ collects the reflected signal with larger strength, while $S_{xx}$, $S_{xy}$, and $S_{yx}$ collect the reflected strength with smaller strength. At 90°, $S_{yy}$ has the maximum strength as the object direction is parallel to the y-direction. The reflected signal collected by $S_{xx}$ is distinguishable but with a smaller strength. $S_{xy}$ and $S_{yx}$ do not contain object reflection. When $\varphi$ increases from 90° to 180°, the variations of signal strengths collected by $S_{xx}, S_{xy}, S_{yx}$, and $S_{yy}$ are opposite to the variations when $\varphi$ increases from 0° to 90°. For $\varphi$ in the range of [90°,180°), the co-polarized components $S_{xx}$ and $S_{yy}$ are identical with the case of $(180° − \varphi)$, such as shown in the $\varphi = 135°$ and $\varphi = 45°$ cases, but the cross-polarized components

$S_{xy}$ and $S_{yx}$ are out of phase in the two cases, which can be used to distinguish the two scenarios.

2) Reflection shape. When $\varphi$ is 0°, the reflection from the root is a horizontal line as the distances between TX/RX to the root are the same at all data collection points. When $\varphi$ changes from 0° to 90°, the reflection curve gradually changes from a flat pattern to a hyperbolic pattern, and the situation reverses when $\varphi$ changes from 90° to 180°.

Therefore, the information of $\varphi$ is carried in the differences of the strength of the signal collected by different polarimetric components and the shape of the reflection signature.

**Vertical inclination angle $\theta$.** Fig. 2(e) shows the radargrams of a root with $\theta$ at -30°, -15°, 0°, 15°, and 30°, and $\varphi$ fixed at 45°. $\varphi$ = 45° is chosen because all the four scattering components are distinguishable, as indicated in Fig. 2(d). As shown in Fig. 2(e), different vertical inclination angles affect the shape and strength of the root reflection signature.

When $\theta$ increases from 0 to 30/−30°, the hyperbolic shape representing the root reflection gradually changes from symmetric to asymmetric, and the asymmetry becomes more obvious with a larger inclination angle. The asymmetry of the reflection shape formed by the negative angles and positive angles appears in the opposite direction because the root's end that is close to the soil surface in these two cases is in the opposite direction. In addition, the inclination angle affects the amplitude distribution in the hyperbolic portion. The maximum reflection intensity is weaker with a larger inclination angle because a part of the EM waves is depolarized to the $z$-direction, resulting in less $x$- and $y$-polarized signals to be collected by the receiver.

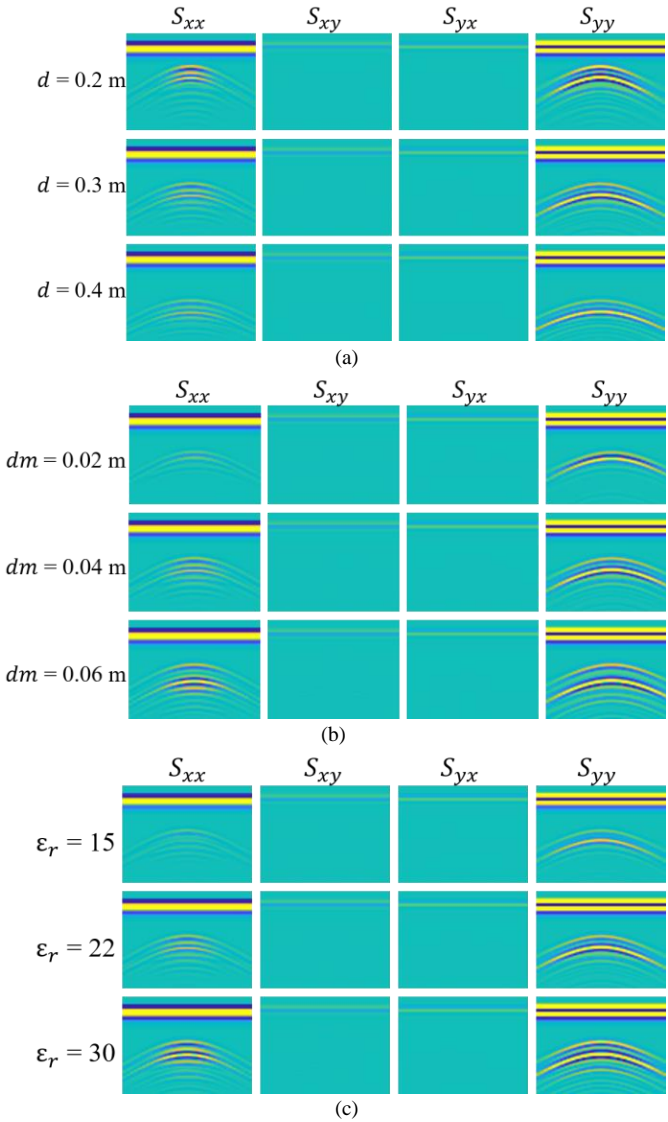

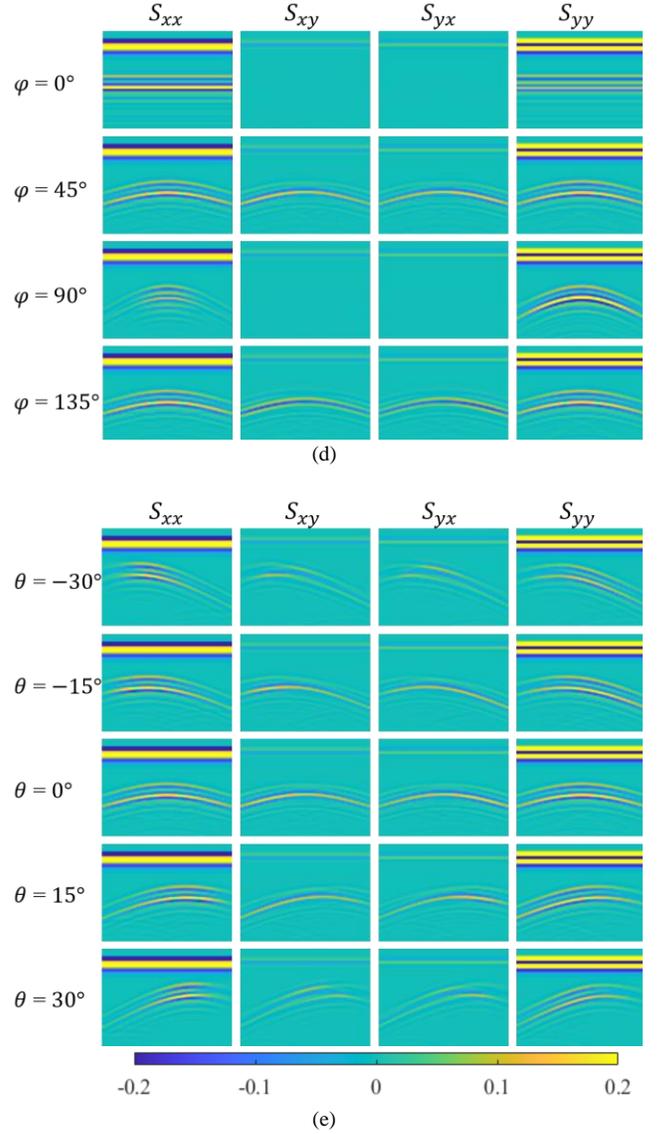

Fig. 2. Radargrams of the different scattering components $S_{xx}$, $S_{xy}$, $S_{yx}$, $S_{yy}$ for a root with different (a) depth $d$, (b) diameter $dm$, (c) relative permittivity $\varepsilon_r$, (d) horizontal angle $\varphi$, and (e) vertical angle $\theta$. Radargrams are represented by colormaps, where different color indicates different values as shown in Fig. 2(e). The signals reflected from the root with different parameters have different characteristics in the multi-polarimetric radargrams, such as intensities, shapes, and spatial locations. These discriminative features in radargrams are useful for parameter estimation.

**Soil heterogeneity.** The heterogeneous soil implemented in gprMax is realized using the semiempirical model proposed in [39], [40] combined with the fractal model described in [41]. The soil in the study has a sand fraction of 0.9, clay fraction of 0.1, bulk density of 1.45 g/cm$^3$, and sand particle density of 1.55 g/cm$^3$. The bulk density of soil and sand is set based on the properties of sandy loam that are ideal for the root growth [42]. 50 different materials with water volumetric fractions (WVF) ranging from 1% to 10% are stochastically distributed with a fractal dimension of 1.5. Different heterogeneous soil scenarios are modeled by generating different material distributions. The properties of the root in the soil are $d = 0.3$ m, $dm = 0.04$ m, $\varepsilon_r = 15$, $\varphi = 45°$, and $\theta = 15°$. The multi-polarimetric radargrams under different heterogeneous soil conditions are shown in Fig. 3. The radargrams in a homogeneous soil environment are also added for comparison.

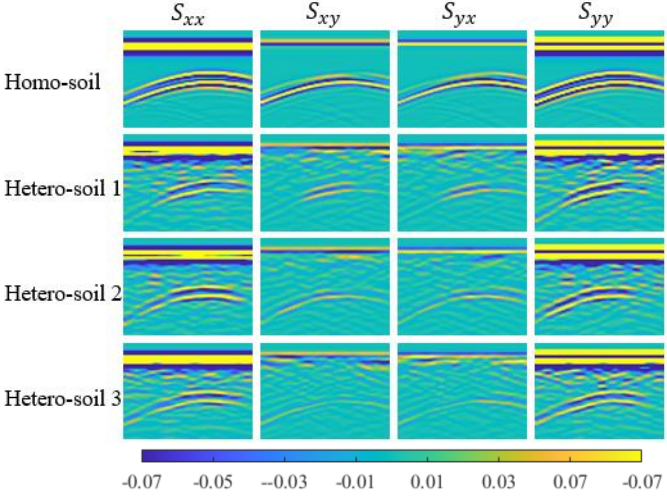

Fig. 3. Comparison of multi-polarimetric radargrams of a root in homogeneous soil (homo-soil), and in different heterogeneous soil (hetero-soil 1-3) environments. The soil heterogeneity introduces clutters and affects the root reflection shape and strength, making it harder to extract effective information to estimate root-related parameters.

As shown in Fig. 3, compared to B-scans under homogeneous soil condition, B-scans under heterogeneous soil conditions not only have much more pronounced clutters, but also show distorted object reflection shapes due to different soil properties at different locations. Moreover, different heterogeneous soil conditions result in different clutter distributions and root reflection shapes and strengths. The unwanted clutters and variation of root reflection shape introduce difficulties in extracting informative features from radargrams to estimate the root parameters. Therefore, it is necessary to eliminate environmental clutter and focus on the root reflection to achieve accurate parameter estimation.

From the above analysis, it can be concluded that multiple root parameters are closely coupled and jointly contribute to the shape, strength, and occupied area of the root reflection signature. Besides, the soil heterogeneity brings in additional environmental clutters and slightly modifies the root reflection shape and strength in radargrams. A conceptual model between multi-polarimetric radargrams and root parameters and soil properties can be established as

$$\begin{bmatrix} S_{xx} & S_{xy} \\ S_{yx} & S_{yy} \end{bmatrix} = f(d, dm, \varepsilon_r, \varphi, \theta, soil). \quad (2)$$

In this work, we aim to build a deep neural network model to learn the relationship between multi-polarimetric radargrams and root parameters while accounting for the soil heterogeneity. By doing so, an accurate estimation of root parameters can be achieved based on multi-polarimetric radargrams.

## III. METHODOLOGY

To effectively estimate the root parameters under heterogeneous soil conditions, we propose the MMI-Net to extract informative features from multi-polarimetric radargrams and build their relationship with the root parameters. Given multi-polarimetric radargrams, the proposed MMI-Net can automatically and simultaneously estimate the five key root-related parameters, i.e., the depth, diameter, relative permittivity, horizontal orientation angle and vertical inclination angle.

The overview network structure of MMI-Net is shown in Fig. 4. The MMI-Net comprises two sub-networks: a MaskNet to generate a mask that highlights the region of the root reflection and a ParaNet to extract the informative features from multi-polarimetric components under the guidance of the mask for accurate parameter estimation. The details of the two sub-networks are described as follows.

### A. MaskNet

The network architecture of the MaskNet is shown in Fig. 4. It takes the concatenated scattering components {$S_{xx}$, $S_{xy}$, $S_{yx}$, $S_{yy}$} along the channel dimension as inputs. The inputs are fed to an encoder-decoder network structure to learn a mask in a pixel-to-pixel manner. The mask highlights the regions of the root reflection signature, which contains most of the information for root parameter estimation. The predicted mask is used to guide the subsequent ParaNet to concentrate on the object reflection region for parameter estimation.

In the encoder branch, three convolution blocks are used to extract the hierarchical encoder features of multiple scattering components. Each convolution block consists of two convolutional layers and one self-calibrated convolution layer, in which each convolutional layer is followed by the rectified linear (ReLU) activation function. Each convolution block is followed by a 2× downsampling operation. In addition, the self-calibrated convolution (SC-Conv) layer [43] is adopted to expand the field of view of the convolutional layer by modeling the long-range spatial and inter-channel dependencies around each spatial location. Such characteristics are significant for extracting the information of the root reflection region from multiple polarimetric inputs. It also avoids contaminating information from irrelevant regions far away from the spatial location.

The framework of the SC-Conv is shown in Fig. 5. For given input feature maps $I$, it is first transformed into two groups of feature maps $I1$ and $I2$ via two individual convolutional layers followed by the LeakyReLU activation function. $I1$ and $I2$ are sent into separate pathways to collect features of different spatial resolutions. For $I1$, a self-calibration operation is carried out with convolution operations including $F1$, $F2$, and $F3$. In the upper path, $I1$ first goes through an average pooling with filter size 4 × 4 and stride 4, then performs feature transformation, and finally it is upsampled to the original resolution. The process can be expressed as

$$I1' = Up\left(F1(AvgPool(I1))\right), \quad (3)$$

where $Up(\cdot)$ is the bilinear interpolation operation that enlarges the extracted features to the original resolution, and $F1(\cdot)$ is the convolution operation. The convolution transformation of low-resolution features obtains a larger field of view. $I1'$ and $I1$ serve as an attention map to guide the feature transformation of $F2(I1)$, namely the calibration process, which is expressed as

$$O1' = F2(I1) \otimes \sigma(I1' + I1), \quad (4)$$

where $\sigma$ is the Sigmoid activation function and $\otimes$ is the element-wise multiplication. $\sigma(I1' + I1)$ is the calibration weights. The final output of $I1$ is

$$O1 = LeakyRelu\left(F3(O1')\right). \quad (5)$$

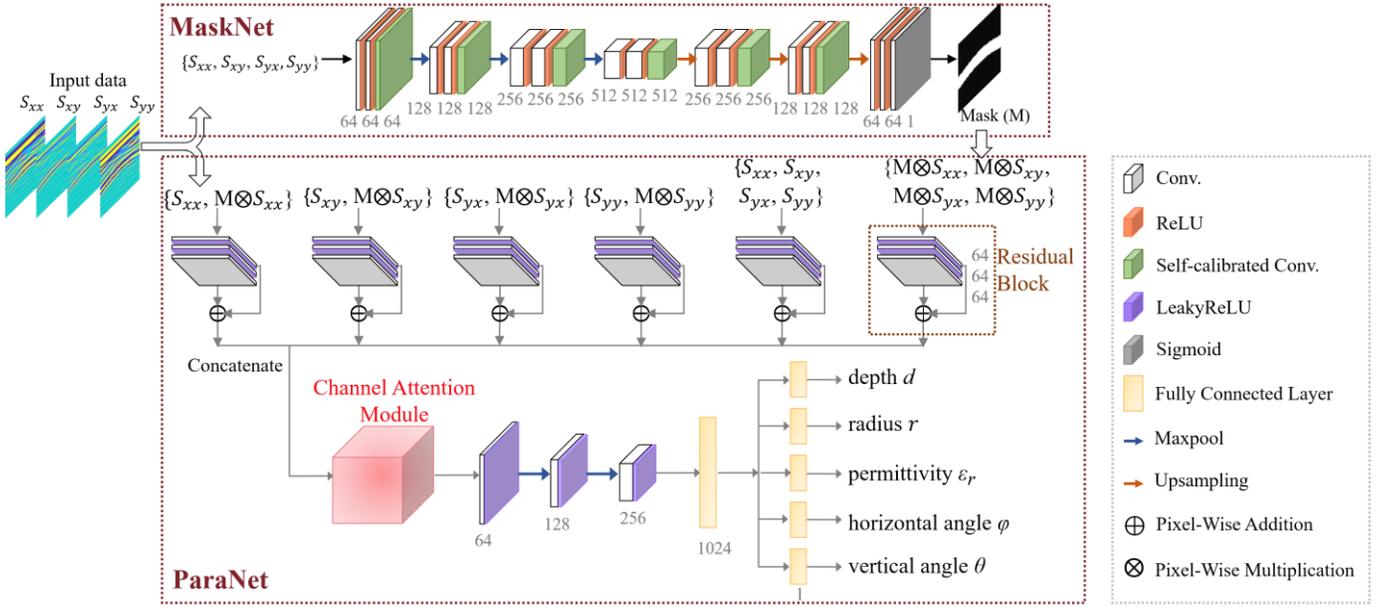

Fig. 4. The overview framework of the MMI-Net. It is composed of a MaskNet to generate a mask that highlights the spatial reflection region of the root reflection, and a ParaNet to simultaneously estimate multiple root-related parameters. In the MMI-Net, each convolutional layer has a kernel size of 3×3 and stride 1. The numbers shown in the figure indicate the numbers of output feature channels (nodes).

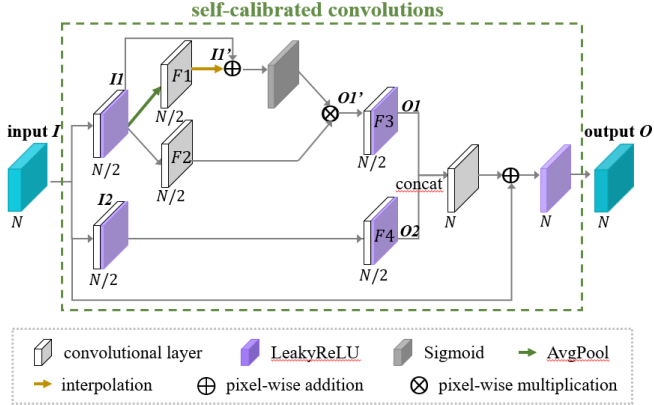

Fig. 5. The framework of self-calibrated convolution. The input features are transformed into two groups of feature maps, i.e., $I1$ and $I2$, which are sent into separate pathways to collect features with different spatial resolutions. The calibration is carried out by using features with a larger fields-of-view as a reference to guide the feature transformation in the original feature space. By fusing informative features in different spatial scales, the network yields better representational capability with a larger receptive field that is essential for the integration of information in multiple scattering components.

For $I2$, a simple convolution operation is performed to preserve the original information of input I, which is expressed as

$$O2 = LeakyRelu\left(F4(I1)\right). \quad (6)$$

The outputs $O1$ and $O2$ are concatenated along the channel direction, then go through a convolutional layer. After that, the concatenated output is added to the input $I$, followed by the LeakyReLU activation function to produce the final output features. By fusing informative features in different spatial scales in the SC-Conv, the output has more discriminative feature representations and achieves a larger receptive field.

Similar to the encoder branch, the decoder branch also includes three convolutional blocks. After each convolutional block, the 2× upsampling operation (bilinear interpolation) is used to enlarge the resolution of features. To estimate the mask, a head that consists of three convolutional layers is used. At last, the MaskNet output the mask with a size of $1 \times H \times W$, where 1 is the output channel number, $H$ and $W$ are the height and width of the mask, respectively.

### B. ParaNet

The estimated mask generated from the MaskNet is taken as an attention map to highlight the RoI of each scattering component. It helps to reduce the interference of environmental clutters in radargrams and guides the ParaNet to pay more attention to the masked region that represents the root reflection signature. To obtain useful information of soil condition from the background and avoid information loss due to potential inaccuracy of the predicted mask, the original scattering components are preserved as part of the inputs. The multi-polarimetric components are employed as inputs as they contain complementary information of the root, which significantly improves the GPR detection capability. Thus, the inputs of ParaNet are the concatenations of original scattering components and the corresponding masked components, as shown in the ParaNet in Fig. 4.

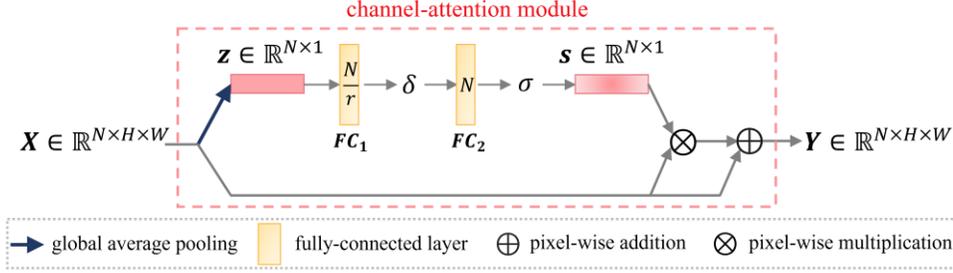

Fig. 6. The schematic illustration of the channel-attention (CA) module. The CA module recalibrates features using global information. The channel weights $s$ are computed by passing the input features $X$ into global average pooling and fully-connected layers. They rescale the input features $X$ to the output features $Y$ to emphasize informative features and suppress redundant features.

The six sets of inputs are independently forwarded to residual blocks [44] to extract features. The residual block contains three convolutional layers where the first two layers are followed by the LeakyReLU activation function. By doing so, six sets of features with both independent and joint information are obtained. These abundant features are fed to a channel attention (CA) module [45] to use global information to selectively emphasize informative features and suppress redundant ones for parameter estimation. Such an attention mechanism adaptively recalibrates channel-wise feature responses by explicitly modelling interdependencies among channels. Thus, it is important for the proposed ParaNet to build the relationship among features of different scattering components.

The schematic illustration of the channel-attention block is shown in Fig. 6. Assuming the input features are $X \in \mathbb{R}^{N \times H \times W}$, where $N$ is the number of features, and $H$ and $W$ are the height and width of spatial dimensions, respectively. The global average pooling is firstly performed on $X$, leading to a channel descriptor $z \in \mathbb{R}^{N \times 1}$, which is an embedded global distribution of channel-wise feature responses. The $k^{th}$ entry of $z$ can be expressed as

$$z_k = \frac{1}{H \times W} \sum_{i}^{H} \sum_{j}^{W} X_k(i,j), \quad (7)$$

where $k \in [1, N]$. To fully capture channel-wise dependencies, a self-gating mechanism is used to produce a collection of per-channel modulation weights $s \in \mathbb{R}^{N \times 1}$:

$$s = \sigma(FC_2 * \delta(FC_1 * z)), \quad (8)$$

where $\sigma(\cdot)$ and $\delta(\cdot)$ represent the Sigmoid and the ReLu activation functions, respectively, $*$ is the convolution operation, and $FC_1$ and $FC_2$ are two fully-connected layers with the output channel numbers of $N/r$ and $N$, respectively. $r$ is set to 16 by default to reduce the computational cost. At last, the weights $s$ are applied to input features $X$ to generate rescaled features $Y \in \mathbb{R}^{N \times H \times W}$. To avoid gradient vanishing problem and keep good properties of original features, the channel-attention weights are treated in a residual fashion, which is expressed as

$$Y = X \otimes s \oplus X \quad (9)$$

where $\otimes$ and $\oplus$ denote the pixel-wise addition and pixel-wise multiplication, respectively. By introducing the CA module, the informative features closely related to the root parameters are emphasized, and the relationship between features in different polarimetric radargrams is modeled.

After the CA module, these highlighted features go through three convolutional layers followed by the LeakyReLU activation function and 2× downsampling operation and then are fed to a fully-connected layer. To simultaneously estimate multiple root-related parameters, the output of the fully-connected layer is followed by a multi-task head where each fully-connected layer corresponds to the estimation of one root parameter.

*C. Loss Function*

To train the MMI-Net, a multi-task loss function is designed, including a mask prediction loss and estimation losses of five root parameters. It can be expressed as

$$L = W_{mask} \times L_{mask} + L_d + L_{dm} + L_\varepsilon + L_\varphi + L_\theta, \quad (10)$$

where $W_{mask}$ represents the weight of mask prediction loss, which is smaller than other losses as the focus of this work is to estimate the root parameters where mask only provides auxiliary guidance for spatial concentration. In this work, it is set to be 0.5. The mask prediction loss $L_{mask}$ is the average binary cross-entropy loss, which is expressed as

$$L_{mask} = -\frac{1}{M^2} \sum_{i=1}^{M} \sum_{j=1}^{M} \bigl(t_{ij} \log(p_{ij}) + (1 - t_{ij}) \log(1 - p_{ij})\bigr), (11)$$

where $t$ and $p$ are the ground truth mask and the predicted mask with a resolution of $M \times M$, respectively. The depth loss ($L_d$), diameter loss ($L_{dm}$), permittivity loss ($L_\varepsilon$), horizontal angle loss ($L_\varphi$), and vertical angle loss ($L_\theta$) are the squared error between the estimated value and the corresponding ground truth. When training in a batch, the loss function is the average $L$ of all training samples in the batch.

## IV. EXPERIMENTS OF THE MMI-NET WITH SIMULATED GPR DATASET

*A. Simulated Dataset Preparation*

To implement the proposed MMI-Net, 3850 sets of data are generated for a root with different parameters in heterogeneous soil using gprMax. Each set of data include the B-scans of $S_{xx}$, $S_{xy}$, $S_{yx}$, $S_{yy}$, and the corresponding root parameters.

The scenario is similar to the one shown in Fig. 1 but on a smaller domain with dimensions of 0.55×0.55×0.85 m³. The soil has a sand fraction of 0.9, a clay fraction of 0.1, a soil bulk density of 1.45 g/cm³, and a sand density of 1.55 g/cm³. The soil is composed of 50 different soil fractures with WVF of 1%-10% and fractal dimension of 1.5. 10 heterogeneous soil conditions are generated with different distributions of soil fractures. The root has its diameter $dm \in [0.02\text{ m}, 0.06\text{ m}]$, relative permittivity $\varepsilon_r \in [15, 30]$, horizontal angle $\varphi \in [0, 179°]$, and vertical angle $\theta \in [-35°, 35°]$, and is buried at a depth $d \in [0.2\text{ m}, 0.4\text{ m}]$. The diameter range is chosen to simulate the coarse root [3]. The permittivity range is for a root with more than 50% water content to guarantee a visible root reflection for analysis in accordance with [38]. The vertical angle range is chosen as most roots have an inclination angle in this range, and beyond this range, the strength of the signal reflected from a root is relatively weak to be detected [4], [46]. The root parameters are all randomly chosen in their corresponding ranges to guarantee diverse scenarios. Orthogonally polarized ($x$- and $y$- polarized) source operating at a center frequency of 1.0 GHz and probes are used as transmitter and receiver. They are moved along $x$-direction with a step size of 2.0 cm. 19 A-scans are collected to form a B-scan. Other setup parameters are the same as those provided in Section II.

*B. Implementation Details of the MMI-Net*

Among the 3850 sets of data, the training and testing data sets are split by a 10:1 ratio. Masks are generated using LabelMe [47] for the 3500 sets of training data. Several examples of masks together with the multi-polarimetric B-scans are shown in Fig. 7. The 3500 sets of training data are augmented to 7000 sets by horizontally flipping the B-scans. The corresponding ground truth of vertical angle $\theta$ are changed to $-\theta$ for the augmented data based on the data collection process. All the images are resized to a size of 64×64 and normalized into a range of [0, 1].

It is noted that more diverse sets of data help to avoid the overfitting of neural networks in the learning process and also improve neural networks' generalization capability. However, generating or collecting a large number of data is certainly time-consuming since each set of data takes 30 mins to be generated using the GPU-accelerated gprMax [48]. In this study, using 3500 sets of training data can already achieve a high estimation accuracy for each parameter. Therefore, to achieve an acceptable estimation accuracy while balancing the time consumed on generating the dataset, 3850 sets of data were used to train and test our algorithm. The 10:1 ratio of training data to testing data is commonly used to validate the performance of deep learning-based methods.

The MMI-Net is implemented with PyTorch on an NVIDIA 2080Ti GPU. End-to-end training is performed based on the loss function presented in Eq. (10). ADAM optimizer with the default parameters is used to optimize the network parameters with an initial learning rate of 0.0001 and a batch size of 70.

*C. Experimental Results of the Estimation Accuracy*

The well-trained MMI-Net is used to estimate the root parameters of 350 testing data. The histograms of estimation errors of root parameters $d$, $dm$, $\varepsilon_r$, $\varphi$, and $\theta$ are presented in Fig. 8. The errors shown in Fig. 8 are calculated as the estimated value minus the true value. As illustrated in Figs. 8(a), the error of depth $d$ has a median of 0.02 cm and a standard deviation of 0.42 cm. 96% of testing data have the error in $d$ within $\pm 1.0$ cm. Fig. 8(b) shows that the estimated error of diameter $dm$ has a median of $-0.0071$ cm and a standard deviation of 0.18 cm. 96% of testing data have the error in $dm$ within $\pm 0.4$ cm. Fig. 8(c) shows the estimated error of relative permittivity $\varepsilon_r$. It has a median of $-0.0092$ and a standard deviation of 1.74. 91% of the testing data have the error in $\varepsilon_r$ within $\pm 3$. Figs. 8(d) and 8(e) show the estimation errors of horizontal and vertical angles $\varphi$ and $\theta$. Their median values are $-0.08°$ and $-0.11°$, and standard deviations are 4.61° and 2.34°, respectively. 73% of the estimated $\varphi$ and 97% of the estimated $\theta$ have the error within ±5°. The error in $\varphi$ is larger due to a wider range of the horizontal angle than the vertical angle in our dataset.

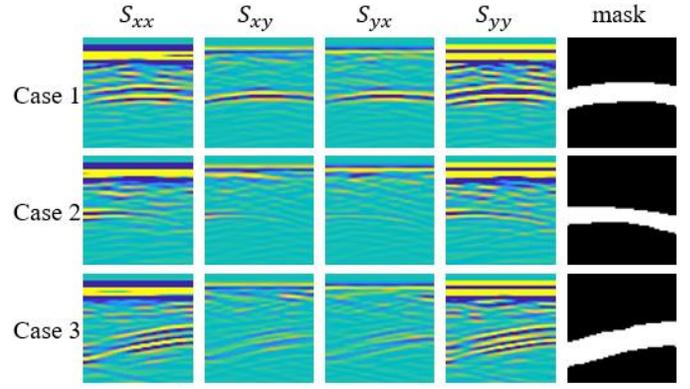

Fig. 7. Multi-polarimetric B-scans and the corresponding masks in three cases.

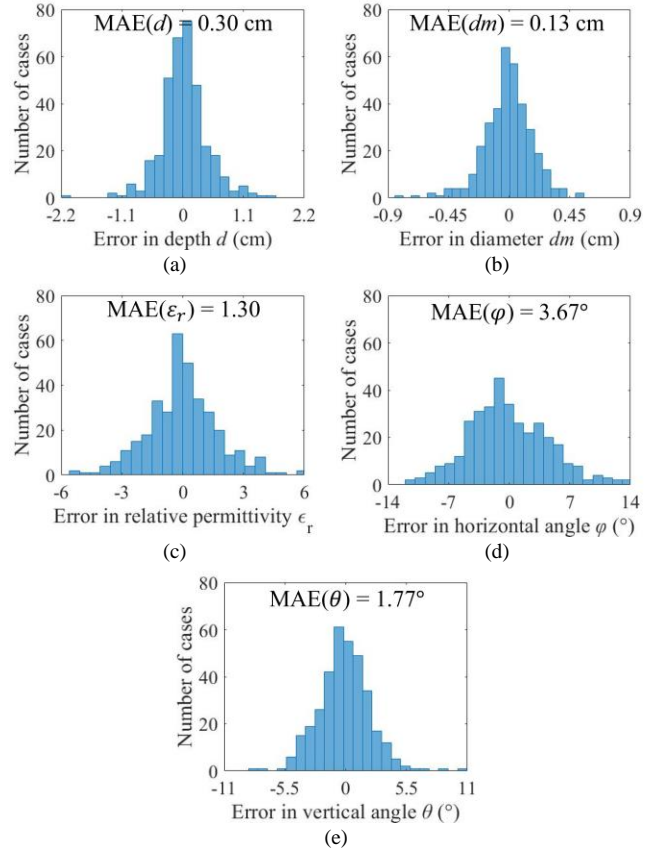

Fig. 8. The histograms of estimation errors of root-related parameters (a) $d$, (b) $dm$, (c) $\varepsilon_r$, (d) $\varphi$, and (e) $\theta$. The mean absolute errors in the 350 testing data are also shown in the figures.

TABLE I
COMPARISON OF PARAMETER ESTIMATION ACCURACY IN THE ABLATION STUDY

| | Effect of mask | | Effect of multi-polarimetric components | | | | Effect of CA | Final |
|---|---|---|---|---|---|---|---|---|
| Network models | A | B | C | D | E | F | G | MMI-Net |
| Original images | √ | | √ | √ | √ | √ | √ | √ |
| Masked images | | √ | √ | √ | √ | √ | √ | √ |
| $S_{xx}$ | √ | √ | √ | | | | √ | √ |
| $S_{xy}$ | √ | √ | | √ | | | √ | √ |
| $S_{yx}$ | √ | √ | | | √ | | √ | √ |
| $S_{yy}$ | √ | √ | | | | √ | √ | √ |
| CA module | √ | √ | √ | √ | √ | √ | | √ |
| MAE($d$) | 0.47 cm | 0.74 cm | 0.55 cm | 0.51 cm | 0.52 cm | 0.53 cm | 0.40 cm | **0.30 cm** |
| MAE($dm$) | 0.22 cm | 0.16 cm | 0.20 cm | 0.20 cm | 0.20 cm | 0.20 cm | 0.16 cm | **0.13 cm** |
| MAE($\varepsilon_r$) | 1.47 | 1.53 | 2.43 | 2.48 | 2.50 | 2.50 | 1.52 | **1.30** |
| MAE($\varphi$) | 5.97° | 5.71° | 38.89° | 7.80° | 8.08° | 38.30° | 6.35° | **3.67°** |
| MAE($\theta$) | 2.68° | 3.10° | 12.02° | 2.72° | 2.88° | 11.75° | 2.40° | **1.77°** |

In addition, the mean absolute errors (MAEs) of $d$, $dm$, $\varepsilon_r$, $\varphi$, and $\theta$ in 350 testing data are 0.30 cm, 0.13 cm, 1.30, 3.67°, and 1.77°, respectively. Considering their different ranges, the averaged errors in percentage (AEP) defined as MAE divided by the full ranges of parameters are 1.50%, 3.25%, 8.67%, 2.05%, and 2.53% for $d$, $dm$, $\varepsilon_r$, $\varphi$, and $\theta$, respectively. Overall, the low MAEs and AEPs on the testing data prove the network's capability in simultaneously and accurately estimating all parameters.

*D. Ablation Study*

Ablation studies are conducted to demonstrate the effectiveness of the network design. The effects of the guidance of mask, the utilization of multi-polarimetric components, and the channel attention module on the parameter estimation accuracy are investigated. The ablated models are retrained while keeping the same settings as the final MMI-Net except for the ablated parts. Table I lists the quantitative comparisons of the estimation accuracy averaged on the testing set.

Model A and Model B are the ablated models with only original images as input and with only masked images as input, respectively. As presented in Table I, Model A and Model B both result in larger MAEs in all parameters compared to the final MMI-Net. This is because without masked images as guidance to concentrate on the RoI [Model A], the network cannot precisely learn the features related to the root parameters in informative regions, thus leads to a large estimation error. However, if only use masked images but without the original images as input [Model B], the information of soil heterogeneity carried in background environments is lost, and a slight inaccuracy of the predicted mask can result in an increased estimation error. This comparison verifies that concatenating the masked images and the original images helps to guide the ParaNet to concentrate on the RoI while taking useful environmental information into account, which significantly improves the estimation accuracy of the network.

Models C-F are the ablated models with only one polarimetric component as input. As presented in Table I, Models C-F have a large error in estimating $d$, $dm$, and $\varepsilon_r$, which is due to the limited detection capability of a single polarized component. For the orientation estimation, a network using only the $S_{xx}$ or $S_{yy}$ component as input yields high estimation error, as using these two components cannot allow for the differentiation of the combination of angles ($\varphi_i$, $\theta_i$) from (180°−$\varphi_i$, 180°−$\theta_i$). A network using only the $S_{xy}$ or $S_{yx}$ component improves the estimation accuracy as the cross-polarized components contain the phase information to differentiate the two intervals of $\varphi$. However, the improvement is constrained by the limited detection capability of cross-polarized components. By using the four polarimetric components as inputs in the final MMI-Net, the complementary information carried in different polarimetric components is integrated, which boosts the network capabilities in accurately estimating the root parameters.

Model G is the ablated model without the CA module. In comparison to MAEs of the final MMI-Net, the estimation accuracy of Model G drops in all parameters, especially in $\varphi$. It can be observed from the result that the CA module can build the interdependencies between features in multi-polarimetric components, which is critical for the estimation of $\varphi$ that has its information carried in the relationship between different polarimetric features. Moreover, the CA module selectively emphasizes the informative features while suppressing the less useful ones, which allows the network to sensitively distinguish useful features for parameter estimation.

By using the masked and original multi-polarimetric images as input, extracting their discriminative features, emphasizing the informative features while suppressing the redundant ones, our proposed MMI-Net successfully builds the relationship between multi-polarimetric radargrams and five key root-related parameters under heterogeneous soil conditions and achieves high estimation accuracy of these parameters.

*E. Generalization Capability of MMI-Net for Roots with Bark*

As coarse roots normally have bark, the effect of the bark in the estimation accuracy of the network is also examined. The bark is a low-permittivity layer with a relative permittivity of 5 and conductivity of 0 [49]. Different ratios between the bark thickness and root radius (bark-to-root ratio) are considered, ranging from 0% (without bark) to 20%. For each bark-to-root ratio, 40 sets of data for a root with different depths, diameters, permittivity, horizontal and vertical orientation angles are generated to test the performance of the MMI-Net. The MMI-Net is trained using the dataset without bark as described in Section IV.B. The mean absolute errors of the five estimated parameters with different bark-to-root ratios are calculated, as listed in Table II. As can be seen, the existence of bark only slightly affects the performance of the MMI-Net. A larger bark-to-root ratio results in a slightly higher estimation error. The error is more obvious in diameter as the bark occupies the space of the root and the network tends to underestimate the diameter. Nevertheless, the model still maintains high estimation accuracy for roots with thin bark.

TABLE II
COMPARISON OF ESTIMATION RESULTS OF ROOT WITH BARK

| Bark-to-root ratio | 0% | 5% | 10% | 15% | 20% |
|---|---|---|---|---|---|
| MAE($d$) | 0.34 cm | 0.33 cm | 0.35 cm | 0.38 cm | 0.44 cm |
| MAE($dm$) | 0.12 cm | 0.15 cm | 0.22 cm | 0.39 cm | 0.60 cm |
| MAE($\varepsilon_r$) | 1.19 | 1.24 | 1.29 | 1.33 | 1.65 |
| MAE($\varphi$) | 3.55° | 3.42° | 4.19° | 4.66° | 4.84° |
| MAE($\theta$) | 1.66° | 1.71° | 1.69° | 1.70° | 1.76° |

## V. EXPERIMENTS OF THE MMI-NET WITH MEASURED GPR DATASET

The performance of the MMI-Net in estimating multiple root parameters is further examined using field-measured GPR data. The dataset preparation, implementation details of the MMI-Net, and experimental results of the parameter estimation accuracy are described in this section.

*A. Measured Dataset Preparation*

The experimental scenario of collecting multi-polarimetric GPR radargrams of the tree root is illustrated in Fig. 9. The experiment was conducted in an outdoor sandy field in Singapore and lasted for two weeks. The relative permittivity of the field varies from 3 to 8 due to the difference in weather conditions and humidity.

As shown in Fig. 9(a), five roots with different diameters and relative permittivity are selected as root samples. Each of them is buried at six different depths $dm \in [0.014\ m, 0.034\ m]$ with different vertical inclination angles $\theta \in [-30°, 30°]$ and horizontal orientation angle $\varphi = 0°$ in the sandy field. Twenty different horizontal angles $\varphi$ of the root with respect to the GPR antenna are realized by moving the GPR antenna along different scanning traces, as shown in Fig. 9(b).

A stepped-frequency multi-polarimetric GPR system is used in the experiment, as shown in Fig. 9(b). The dual-polarized Vivaldi antenna described in [50] is used as the transmitter (TX) and receiver (RX) in a monostatic setup. The antenna has two ports to transmit/receive horizontally and vertically polarized signals, respectively. The antenna is sealed in a foam box and is surrounded by absorbers to reduce environmental noise. The two ports of the antenna are connected to a vector network analyzer (Keysight VNA P5021A), forming a stepped-frequency GPR system. The system simultaneously collects the four polarimetric components $S_{xx}$, $S_{xy}$, $S_{yx}$, and $S_{yy}$. For each A-scan, the GPR system records 1001 frequency samples in a band ranging from 0.4 GHz to 4.0 GHz. The collected frequency domain data are transformed to the time domain via inverse Fourier transform. 19 A-scans are collected along each scanning trace with a step size of 2 cm and are combined into a B-scan. Mean subtraction [51] is used to pre-process the B-scans, where the subtracted background trace is the average trace of a B-scan collected in the field when no root is buried.

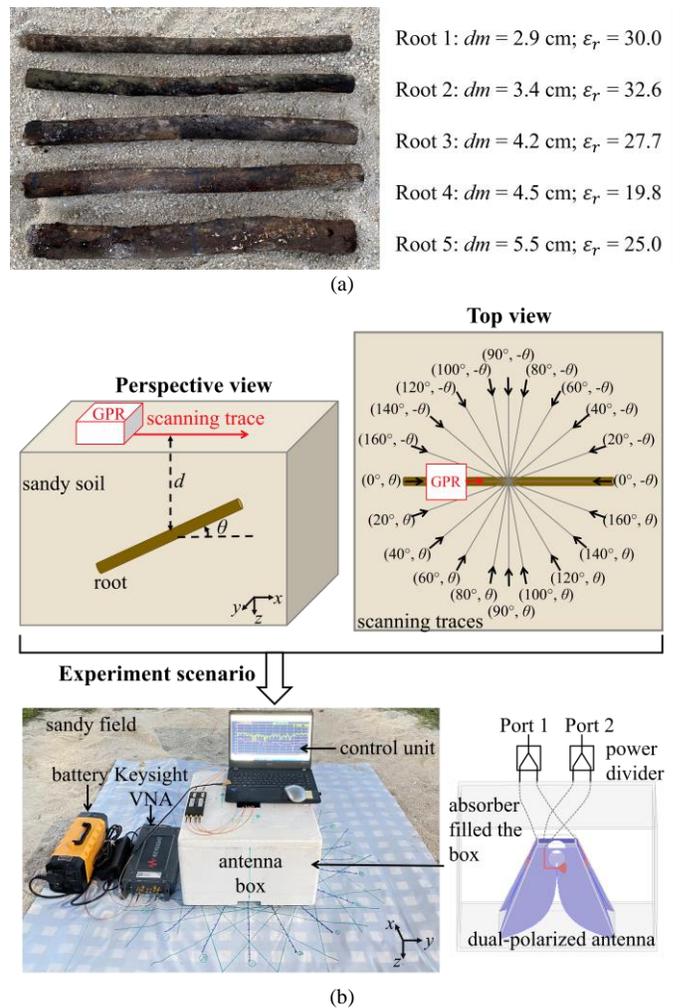

Fig. 9. (a) Five root samples and their corresponding parameters for the collection of experimental data. (b) The schematic view of the experiment scenario and the view of the real scenario. The root is buried at different depths and veritical angles. Different horizontal angles are realized by moving the GPR antenna along different scanning traces. The dual-polarized antenna presented in [50] is used to simultaneously collect multi-polarimetric scattering components $S_{xx}$, $S_{xy}$, $S_{yx}$, and $S_{yy}$. 550 sets of B-scans are collected for the five root samples.

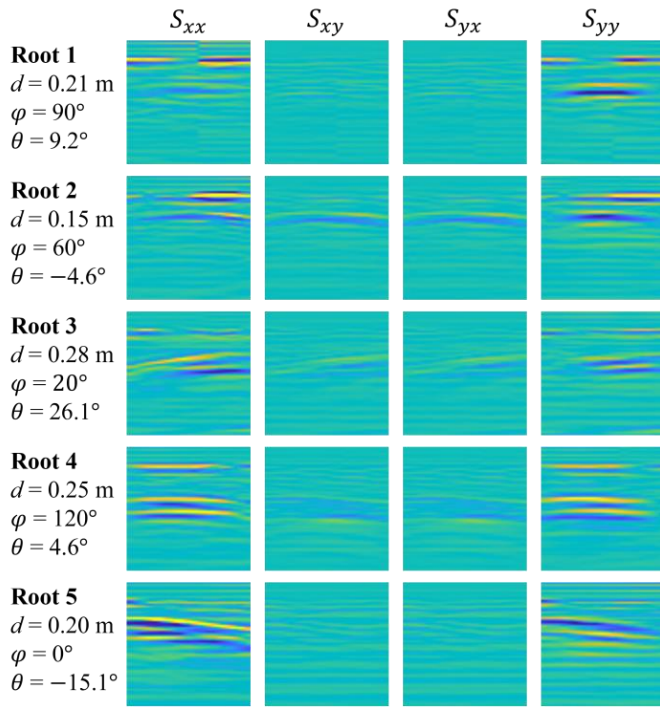

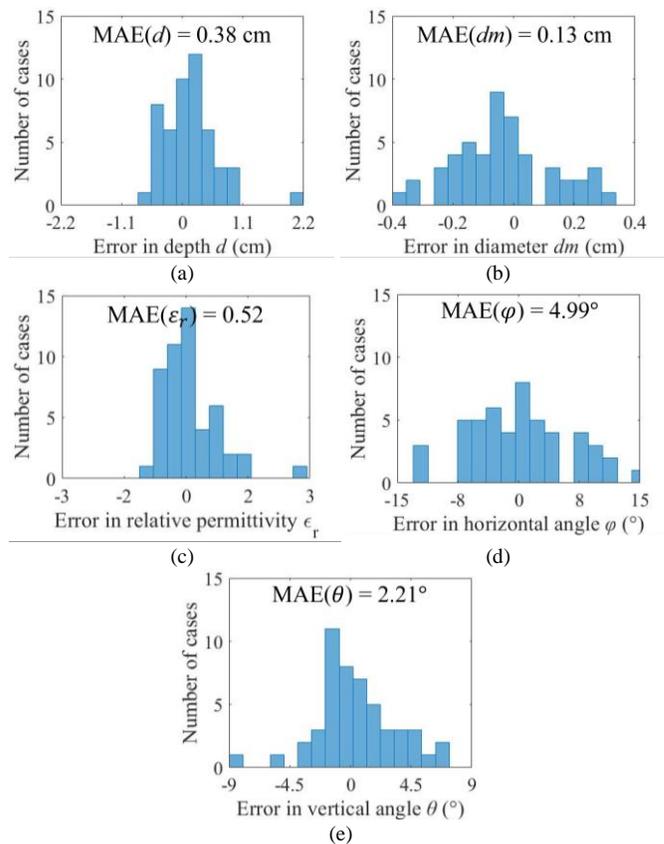

Fig. 10. Illustration of collected B-scans of the five root samples in the sandy field. The root reflection with different characteristics can be observed, and some environmental clutter slightly distorts or disguises the reflected signatures.

Fig. 11. The histograms of the estimation errors of root-related parameters (a) $d$, (b) $dm$, (c) $\varepsilon_r$, (d) $\varphi$, and (e) $\theta$. The mean absolute errors in the 50 testing data are also shown in the figures.

In the experiment, 550 sets of multi-polarimetric B-scans are successfully collected for the five root samples with different depth, horizontal, and vertical angles. Fig. 10 illustrates several collected B-scans of the five roots in different spatial arrangements. The root reflection with different characteristics can be observed, and some environmental clutter slightly distorts or disguises the reflected signatures.

*B. Implementation Details of the MMI-Net*

The 550 sets of B-scans are randomly divided into 500 sets of training data and 50 sets of testing data. All the B-scans are resized to 64×64 and normalized to the range of [0, 1]. The MMI-Net pre-trained by the simulated data (as described in Section IV) is fine-tuned using the 500 sets of the field-measured training data. The purpose of fine-tuning the network is to avoid overfitting. In the fine-tuning process, a smaller learning rate of 0.00001 is used. After training, the network is used to estimate the root parameters of the 50 sets of field-measured testing data.

*C. Experimental Results of the Estimation Accuracy*

The histograms of parameter estimation errors of the fine-tuned MMI-Net are shown in Fig. 11. In Fig. 11(a), the maximum estimation error of the depth $d$ is 2.10 cm and 96% of the testing data have the error within ±1.0 cm. As shown in Fig. 11(b), the estimation error of the diameter $dm$ is distributed within ±0.33 cm. 74% of the testing data have the error within ±0.2 cm. Fig. 11(c) shows that the maximum estimation error of relative permittivity $\varepsilon_r$ is 3.10 and 98% of the testing data have the error in $\varepsilon_r$ within ±2. Figs. 11(d) and 11(e) show the estimation errors of horizontal and vertical angles $\varphi$ and $\theta$. Their maximum estimation errors are 14.72° and 7.32°, respectively. 78% of the estimated $\varphi$ values have the error within ±8°, and 90% of the estimated $\theta$ values have the error within ±5°. The MAEs of $d$, $dm$, $\varepsilon_r$, $\varphi$, and $\theta$ are 0.38 cm, 0.13 cm, 0.52, 4.99°, and 2.21°, respectively.

Compared with the results obtained with the simulated testing data in Section IV, the estimation accuracy for the measured testing data is only slightly degraded for the parameters $d$, $\varphi$, and $\theta$. The decreased accuracy could be caused by the deviation of the scanning traces and deviation of the placement of roots in the measurement. The estimation accuracy of $\varepsilon_r$ in the measured testing data is improved, which is because the root's relative permittivity in the measured dataset is less random than the simulated dataset. Overall, the low MAEs in each parameter achieved on the field-measured testing data demonstrate the applicability of the MMI-Net in accurately estimating multiple root parameters in the field experiments.

## VI. COMPARISON AND DISCUSSION

In this section, the effectiveness of the proposed MMI-Net in root parameter estimation is compared with the performance of the existing methods. The limitations of the proposed method and the future work to address challenges in real-life root surveys are also discussed.

## A. Comparison with Existing Methods

To compare the effectiveness in parameter estimation of the proposed MMI-Net with existing methods, a series of comparative studies is performed on the estimation of root depth, diameter, and orientation angles. The estimation of the root relative permittivity via GPR radargrams has not been investigated before. The only work studying the root relative permittivity was carried out by harvesting roots and measuring their relative permittivity values using the Keysight 85070E dielectric probe [38]. Therefore, the comparison study of the estimation of the root relative permittivity is not performed.

**Root depth $d$.** The performance of the proposed MMI-Net in estimating root depth is compared with the conventional time-velocity method described in [3] and the randomized Hough transform (RHT) method implemented in [5], [52]. In the conventional time-velocity method, the soil relative permittivity is the average value in different soil regions under heterogeneous soil conditions [53].

To compare the estimation accuracy of different methods, 20 sets of data are generated for a root buried at different depths in heterogeneous soil. The diameter and relative permittivity of the root are randomly selected in the pre-defined range, and the root is oriented perpendicularly to the scan trace without any inclination. The estimation accuracy of the time-velocity method [3], RHT [5], and the proposed MMI-Net are listed in Table III. The MMI-Net outperforms the other two methods by a large margin. The high estimation accuracy of the MMI-Net is achieved by effectively taking useful soil heterogeneity information from the background in the estimation process, whereas the time-velocity method fails to do so and the RHT can only roughly estimate the soil permittivity in a region consisting of reflection hyperbola.

TABLE III
COMPARISON OF ESTIMATION RESULTS OF ROOT DEPTH

| Methods | MAE($d$) |
|---|---|
| Time-velocity method [3] | 1.67 cm |
| RHT [5] | 1.45 cm |
| MMI-Net | 0.27 cm |

**Root diameter $dm$.** The conventional methods for root diameter estimation build regression models between the root diameter and GPR waveform indices [6]-[12]. Among all the waveform indices, the sum of time intervals between zero crossings for all reflection waveforms $\Sigma T$ has proven to be well correlated with the root diameter and is less susceptible to other root-related parameters [10]-[12]. Therefore, in this study, the performance of MMI-Net is compared with the accuracy of the regression model built with $\Sigma T$.

As most existing models perform root diameter estimation without considering root orientations, 40 sets of data are generated for a root with randomly selected diameter, depth, and relative permittivity and with $\varphi = 90°$ and $\theta = 0°$ for a fair comparison. Following [11], 20 sets of data are used to build the regression model between diameter and $\Sigma T$, as shown in Fig. 12(a). The regression model is

$$dm = 2.6724 \times \Sigma T - 1.8945. \qquad (12)$$

The model has a coefficient of determination $R^2$ of 0.85 and an estimation MAE of 0.32 cm. The model is further validated using the other 20 sets of data. The estimation result has a $R^2$ of 0.89 and an MAE of 0.29 cm, as shown in Fig. 12(b).

In comparison, the proposed MMI-Net achieves MAEs of 0.14 cm and 0.13 cm on the modeling data and testing data, respectively, which outperforms the regression model [11]. The regression model is built based on the assumption that the diameter is the only factor that affects $\Sigma T$, whereas the soil condition and other root parameters especially the relative permittivity also affect the variation of $\Sigma T$. Compared with the regression model, the MMI-Net learns the relationship between the radargrams with root diameters while considering other root-related parameters and the soil environment, thus achieving a higher estimation accuracy.

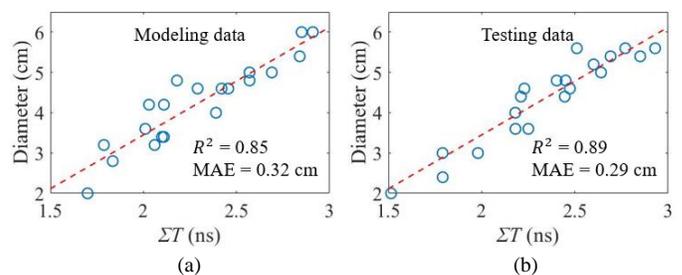

Fig. 12. The regression model [11] for root diameter estimation. (a) The modeling data used to build the regression model between root diameter and $\Sigma T$. (b) The estimated root diameter using the model on the testing data. In comparison, our MMI-Net achieves a lower MAE of 0.14 cm and 0.13 cm on the modeling data and testing data, respectively.

**Root orientation angles.** The estimation accuracy in root orientations of the MMI-Net is compared with the algorithm proposed in [4], and the results are shown in Table IV. The algorithm in [4] builds a mathematical model linking the hyperbolic shape of root reflection with the root orientation angles. Note that the model in [4] is not suitable for the implementation in the entire horizontal angle range, and the data selection along the hyperbola shape was performed manually. Therefore, the estimation results shown in Table IV are the calculated MAEs using the reported results in [4] in a sandy field with a 900-MHz GPR system and a 1600-MHz GPR system.

As can be seen in Table IV, the MMI-Net outperforms the mathematical algorithm in both the horizontal and vertical angles. The large estimation errors in the mathematical model are due to the limited detection capability of a single polarized antenna and the unstable process of the nonlinear fitting. Also, as mentioned in [4], the algorithm does not take the root diameter into account, limiting its applicability for estimating orientation angles for different roots. On the contrary, the proposed MMI-Net is capable of accurately estimating the root orientation angles while considering other root parameters by automatically detecting the root reflection region and effectively extracting discriminative features from multi-polarimetric radargrams.

13TABLE IV
COMPARISON OF ESTIMATION RESULTS OF ROOT ORIENTATION ANGLES

| Methods | MAE($\varphi$) | MAE($\theta$) |
|---|---|---|
| Mathematical algorithm (900-MHz) [4] | 10.44° | 3.87° |
| Mathematical algorithm (1600-MHz) [4] | 8.26° | 4.11° |
| MMI-Net on simulated dataset | 3.67° | 1.77° |
| MMI-Net on measured dataset | 4.99° | 2.21° |

All the comparison results verify that the proposed MMI-Net achieves higher estimation accuracy for each root-related parameter compared with the existing methods. Moreover, in contrast to conventional methods that only estimate one root parameter at a time and ignore the randomness of root orientations, our method is capable of simultaneously estimating multiple root parameters together with its orientation angles. The accurately estimated root-related parameters by the MMI-Net can provide more information for the investigation of roots' ecological function and the reconstruction of root system architectures.

*B. Limitations of the MMI-Net and Future Work*

The proposed MMI-Net has demonstrated its high performance in estimating multiple root-related parameters in heterogeneous soil environments. However, it still has some limitations when it is applied to real-life root surveys.

1) In real-life subsurface environments, other objects (such as rocks and pipes) can be buried in the same area as the tree root. These subsurface objects also produce EM reflections, causing ambiguity of root recognition. The root reflection area needs to be distinguished first before applying the proposed method. The classification of subsurface reflections [54] can be helpful to address this issue.

2) The paper considers the root that buried in heterogeneous soil environment made up of a random distribution of soil fractures with different water volumetric fractions but with fixed sand-clay ratio and density. However, the soil in which a root system exists may be characterized by different compaction levels and has different properties, which could affect the parameter estimation accuracy. A more diverse dataset considering different soil environments and compaction levels can be helpful to address this issue, which will be investigated in our future work.

3) The isolated root is studied in this work as a proof-of-concept of utilizing neural networks to simultaneously estimate multiple root parameters. The practical root system has more intricate structures, and its radargrams may contain overlapping reflection signatures from adjacent roots, which causes difficulties in the data interpretation and root parameter estimation. Therefore, the proposed method should be further improved for parameter estimation in multiple root scenarios.

It is noted that our network is a baseline model for the multi-parameter estimation task. More complex networks can be developed to address the aforementioned limitations in real-life root surveys.

## VII. CONCLUSION

In this paper, a novel neural network architecture, called MMI-Net, is proposed for the estimation of multiple parameters of the root in heterogeneous soil environments. The MMI-Net has two closely coupled sub-networks to predict a mask that highlights the root reflection region and then use the mask to guide the extraction of informative features to estimate multiple parameters. The network is specially designed to alleviate the detrimental effects of background clutter caused by soil heterogeneity and to model the relationship between the root reflection signature with multiple root-related parameters. Experiments using both simulated and measured GPR datasets verify that given multi-polarimetric radargrams, the network is capable of accurately estimating all the root-related parameters under consideration.

To the best of the authors' knowledge, this is the first work that takes into account of both the horizontal and vertical orientation angles of roots while estimating other root parameters. It successfully links the characteristics of GPR multi-polarimetric radargrams with the combined contributions of root-related parameters, and uses this relationship to achieve accurate estimation of multiple root-related parameters. In addition, the idea of combining automatic RoI recognition and parameter estimation using deep neural networks addresses the difficulties in data interpretation for heterogeneous soil. Although targeting at the parameter estimation of the tree root, the proposed method could be extended to characterize elongated subsurface objects in general, including the utilities such as pipes and cables. The idea of utilizing neural networks to extract informative features from radargrams and estimate key parameters of subsurface objects could be further expanded to the classification and imaging of subsurface objects in heterogeneous soil environments.

ACKNOWLEDGMENT

The authors would like to thank the FYP students Miss Yixuan Wu, Mr. Jia Dian Tan, and Mr. Jun Wei Tan for their help in the process of field measurement.